\documentclass[11pt]{article}

\usepackage[preprint]{acl}
\usepackage{times}
\usepackage{latexsym}

\usepackage[T1]{fontenc}
\usepackage{amsmath}
\usepackage{amssymb}
\usepackage{booktabs}
\usepackage{placeins}
\usepackage{flafter}

\usepackage[utf8]{inputenc}

\usepackage{microtype}

\usepackage{inconsolata}

\usepackage{graphicx}

\title{Polarity Composition Hypothesis: Causal Evidence from Negation in NLI}

\author{
  Nhi Hoai Doan \quad\quad Tatsuya Hiraoka \quad\quad Munachiso Samuel Nwadike \quad\quad Kentaro Inui \\
  Mohamed bin Zayed University of Artificial Intelligence (MBZUAI)\\
  RIKEN\\
  \textbf{Correspondence: }\texttt{nhi.doan@mbzuai.ac.ae}
}

\begin{document}
\maketitle
\begin{abstract}
We investigate how language models process negation in natural language inference (NLI). We propose the \textit{Polarity Composition Hypothesis}, a high-level causal model of this computation. Behavioral evaluations show that language-model predictions largely conform to the proposed hypothesis. Using Distributed Alignment Search, we identify causally effective subspaces aligned with the variables in the polarity-composition hypothesis. These representations transfer across negation forms and partially to a subset of MNLI. We release the dataset and code\footnote{\url{https://github.com/hnhine/polarity_composition}}.
\end{abstract}

\section{Introduction}

Large language models (LLMs) achieve strong performance on natural language inference (NLI), where a model determines whether a hypothesis follows from, contradicts, or remains undetermined by a premise \citep{bowman-etal-2015-large, williams-etal-2018-broad}. However, a correct label does not reveal how the model reaches its decision. Models can exploit lexical correlations and annotation artifacts while remaining brittle on controlled contrasts \citep{gururangan-etal-2018-annotation, hossain-etal-2020-analysis}. Negation provides a useful diagnostic because changing a single operator can shift an event from affirmative to negative polarity while largely preserving its underlying content.
\begin{figure*}[ht]
    \centering
    \includegraphics[width=\textwidth]{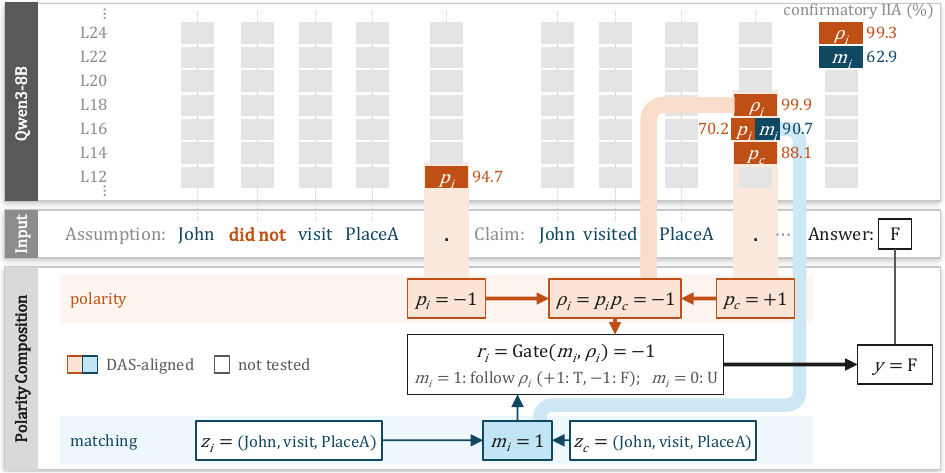}
    \caption{\textbf{Polarity Composition Hypothesis.} \textit{Bottom:} for a claim and assumption events, the hypothesis first finds an event with matching content ($m_i$). If none matches, it predicts \textsc{Unknown}; otherwise, it compares their polarities ($\rho_i = p_i p_c$), predicting \textsc{True} for agreement and \textsc{False} for disagreement. \textit{Top:} coloured cells mark DAS-aligned subspaces of Qwen3-8B hidden states, with confirmatory interchange intervention accuracy~(IIA,~\%; \S\ref{sec:causal-flow}).}
\label{fig:polarity-composition}

\end{figure*}

Natural logic models inference by assigning relations to atomic edits and composing them into a sentence-level relation \citep{maccartney2007natural, maccartney2009extended}. Prior work has used controlled NLI data and interventions to test whether fine-tuned encoders partially implement such algorithms \citep{geiger-etal-2020-neural, geiger-etal-2021-causal, rozanova-etal-2024-estimating}. In a complementary setting, \citet{zhou2026language} study how decoder-only language models process negated expressions, identifying constructive and suppressive mechanisms together with late-layer attention shortcuts. Whereas their work focuses on how negation changes concept representations, we study the negation relation between multiple assumption events and the event in the claim. We take one idea from natural logic: when an assumption and claim share their content, they differ by at most one atomic edit, the negation operator. Solving the task therefore requires separating whether the two sentences describe the same event from whether their polarities agree. It remains unclear whether frozen decoder-only LLMs separate these two components and use them in NLI.

To address this gap, we construct a controlled task that includes multiple assumption events and a single claim. As shown in Figure~\ref{fig:polarity-composition}, we hypothesize that models process negation in NLI  based on two key factors: \textit{event matching} and \textit{relative polarity} influenced by negation. Event matching determines whether an assumption event and the claim describe the same event, while relative polarity indicates whether their polarities are equal.

We use this hypothesis as a high-level causal model and test it in Qwen3-8B and Phi-4 Mini Instruct, two open-weight models that pass our behavioral test, using Distributed Alignment Search (DAS), with all model parameters frozen \citep{geiger-etal-2024-finding}.  Our main contributions are:
\begin{itemize}

    \item We introduce \textit{Polarity Composition Hypothesis}, which proposes an internal mechanism on how LLMs handle negation in NLI(\S\ref{sec:polarity-hypothesis}). 
    
    \item We demonstrate that both models behaviorally follow the predictions of the proposed causal hypothesis (\S\ref{sec:behavioral}).

    \item We identify causally effective representations aligned with the hypothesis and show that they interact in determining the NLI output (\S\ref{sec:causal-flow}).

    \item The learned representations transfer across negation forms and to MNLI examples, although relative relation transfer depends on matching context (\S\ref{sec:robustness}).
\end{itemize}

Together, these results support a high-level causal explanation of how LLMs process negation in NLI: identify whether events match, form a polarity relation for matched events, and use the two signals jointly to solve.

\section{Related Work}
\label{sec:related-work}
 
\paragraph{Negation in NLI} Natural logic performs inference directly over linguistic expressions by assigning relations to atomic edits, projecting them through monotonicity contexts, and composing the results \citep{maccartney2007natural,maccartney2009extended}. This framework has motivated controlled evaluations of monotonicity and negation, including MED, MoNLI, NaN-NLI and ScoNe-NLI \citep{yanaka-etal-2019-neural, geiger-etal-2020-neural, truong-etal-2022-another, she-etal-2023-scone}. These evaluations show that neural NLI models often fail to generalize across lexical, scope, and negation contrasts. Broader studies report similar sensitivity in autoregressive language models and show that negation remains weakly represented in NLI benchmarks \citep{hossain-etal-2020-analysis, truong-etal-2023-language,  garcia-ferrero-etal-2023-dataset,vrabcova2025negation}. We use natural logic as motivation, not as an implementation: we keep only the atomic negation edit and do not model monotonicity projection.  A similar separation of event content from polarity appears in formal semantics: situation semantics describes a basic state of affairs by a relation, its arguments, and a polarity \citep[\S2]{sep-situations-semantics}, and event semantics \citep{Davidson1966-DAVTLF} treats negation as an operator applied to an event description \citep{Champollion2010QuantificationAN, bernard2018negative}. These theories also treat quantifier scope, which our hypothesis does not.  

\paragraph{Causal Analysis} The explicit intermediate structure of natural logic makes NLI a useful setting for causal analysis. An interchange intervention replaces part of a model's hidden state on one input with the state computed on another input, and checks whether the output changes as a high-level causal model predicts. \citet{geiger-etal-2020-neural} combined probing and interchange interventions to test whether MoNLI-fine-tuned BERT partially implements an algorithm linking lexical entailment and negation. This approach, which implements causal inference~\cite{pearl-2001-direct-and-indirect-effects} on neural language models, was extended to compositional NLI by \citet{geiger-etal-2021-causal}. More recently, \citet{rozanova-etal-2023-interventional,rozanova-etal-2024-estimating} intervened on natural-logic features and estimated their effects on NLI predictions. These studies focus on task-trained encoders and the interaction between lexical relations and monotonicity. In a complementary setting, \citet{zhou2026language} identify constructive and suppressive mechanisms for negated concepts in decoder-only LLMs. Neither line of work identifies a mechanism by which LLMs solve NLI involving negation. We address this gap by proposing and causally testing Polarity Composition Hypothesis as candidate mechanism.

\paragraph{Variable Identity} Our analysis across layers and token positions relates to information routing in factual recall and belief tracking \citep{geva-etal-2023-dissecting, prakash2026language}. Learned intervention subspaces raise questions about variable identity and intervention faithfulness: a subspace that changes the output as predicted may carry the source label or another correlated feature instead of the intended variable. Successful patching can activate alternative pathways or produce divergent downstream representations \citep{makelov2024is, wu2024reply, grant2026addressing}, while RAVEL and MIB evaluate the isolation and recovery of causal variables \citep{huang-etal-2024-ravel,pmlr-v267-mueller25a}. We therefore freeze each learned subspace and compare pre-specified identity hypotheses without identity-specific refitting. This test distinguishes among candidate interpretations considered at given site, but does not claim to recover a unique intrinsic variable.

\section{Polarity Composition Hypothesis}
\label{sec:polarity-hypothesis}
\begin{figure}[t]
    \centering
    \includegraphics[
        width=0.9\columnwidth,
        trim=1cm 14.2cm 10.5cm 1.5cm,
        clip
    ]{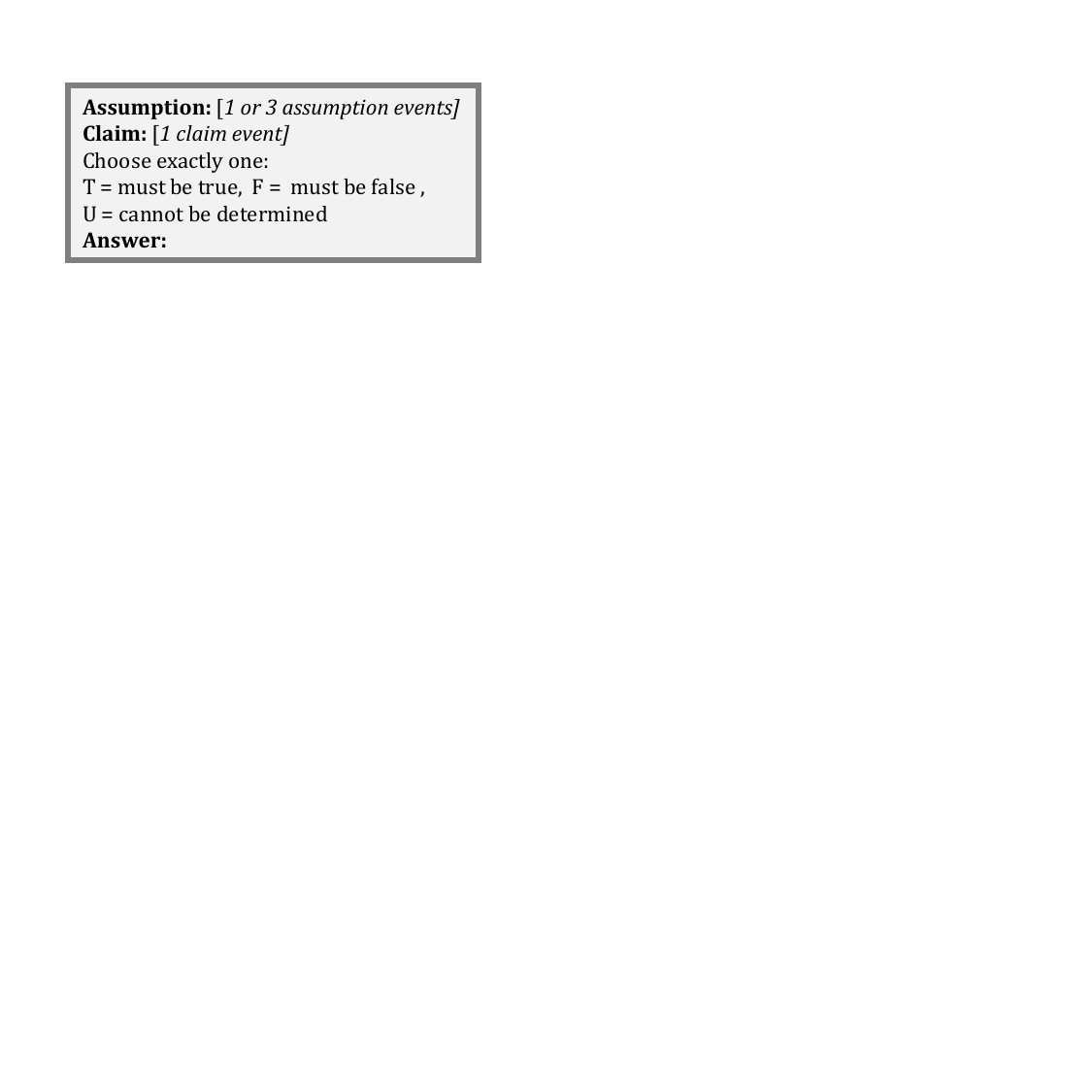}
    \caption{\textbf{Prompt template.}
    Inputs contain one or three assumption events and one claim.}
    \label{fig:prompt-template}
\end{figure}
Negation alone does not determine the inference relation between two sentences. We use \textit{polarity} to indicate whether a sentence
affirms or negates an event description. Consider the following pair:

\begin{quote}
\textbf{Assumption:} John did not visit PlaceA.\\
\textbf{Claim:} John visited PlaceA.
\end{quote}
The sentences share the same polarity-free event content (\textit{John}, \textit{visit}, \textit{PlaceA}) but have opposite polarities because the assumption is negated, yielding \textsc{False}. If the claim is replaced with \textit{John visited PlaceB}, the assumption remains negated, but the event contents no longer match. Its polarity should therefore not affect the output, and the appropriate label is \textsc{Unknown}.

This contrast motivates the \textbf{Polarity Composition Hypothesis}. The hypothesis separates event matching from relative polarity and allows the polarity relation to affect the decision only for a matching event. Polarity agreement yields \textsc{True}, disagreement yields \textsc{False}, and no match yields \textsc{Unknown}. With multiple assumption events, only the relation associated with the matching event should contribute.

We consider inference over assumption events $A=(e_1,\ldots,e_i,\ldots,e_n)$ and a claim event $C=e_c$. Each event is represented as $e_i=(z_i,p_i)$, where $z_i=(s_i,v_i,o_i)$ is its subject-verb-object content and $p_i\in\{-1,+1\}$ is its polarity. Affirmative and negative events take values $+1$ and $-1$, respectively. The claim is represented analogously as $e_c=(z_c,p_c)$. 

We formalize the hypothesis by variables as:

\begin{equation}
\begin{aligned}
    \rho_i
    &= \operatorname{PolComp}(p_i,p_c)
     = p_i p_c,\\
    m_i
    &= \mathbf{1}[z_i=z_c],\\
    r_i
    &= \operatorname{Gate}(m_i,\rho_i)
     = m_i\rho_i,\\
    r(A,C)
    &= \sum_i r_i,
    \qquad
    y(A,C)=g(r(A,C)).
\end{aligned}
\label{eq:causal-flow}
\end{equation}
Here, $\rho_i=+1$ when the two polarities agree and $\rho_i=-1$ when they disagree. Flipping either polarity reverses $\rho_i$, whereas flipping both preserves it. Matching $m_i$ requires identity of subject, verb, and object, which works when the subject and object are proper names.\footnote{Quantified subjects are discussed in the Limitations section.} At the level of the high-level algebra, $\rho_i=p_ip_c$ is defined for any event pair, and $m_i$ determines whether it contributes to the output. This does not require the model to instantiate a transferable relation code when $m_i=0$.

The decision function satisfies $g(+1)=\mathrm{T}$, $g(-1)=\mathrm{F}$, and $g(0)=\mathrm{U}$. Our main setting assumes at most one matching assumption, $\sum_i m_i\leq1$, so that $r(A,C)\in\{-1,0,+1\}$. Multiple matching or mutually contradictory assumptions require an additional aggregation rule, which is outside the present scope.
Figure \ref{fig:polarity-composition} illustrates the hypothesis.

\begin{figure*}[ht]
    \centering
    \includegraphics[width=0.92\textwidth]{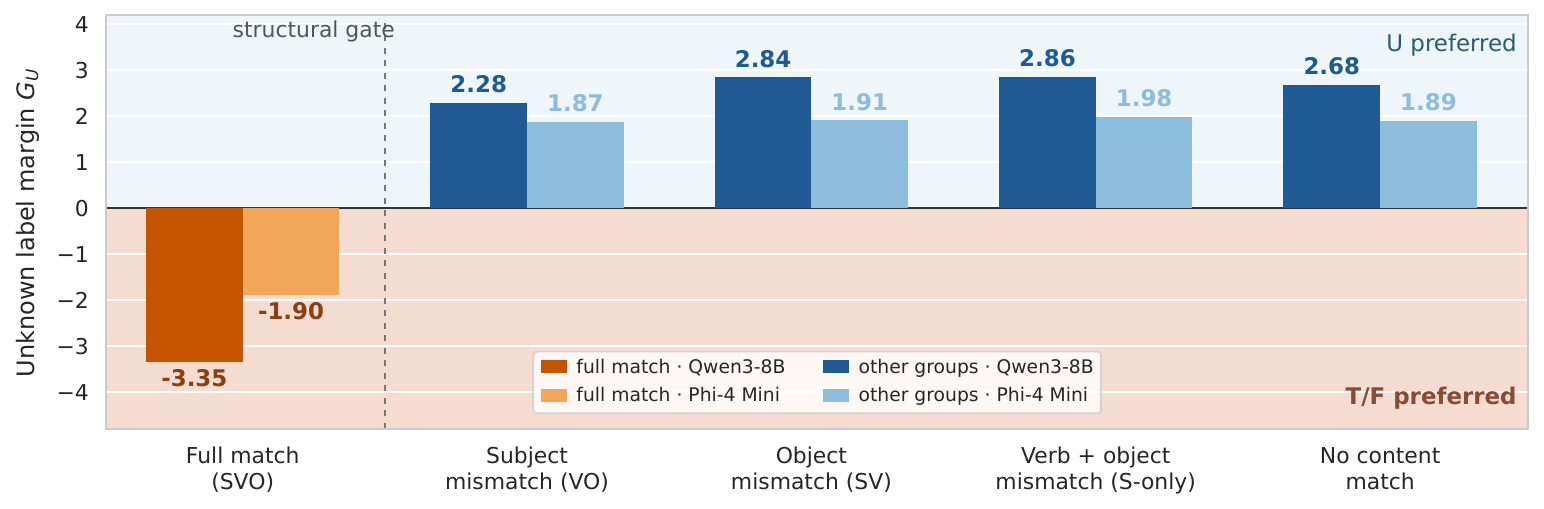}
    \caption{\textbf{Effect of event matching on output preference.}  Both exhibit $G_U<0$ only for full SVO matches, and $G_U>0$ for any mismatch condition, consistent with event matching determining whether polarity affects output.}

    \label{fig:behavioral-gate}
\end{figure*}

Next, \S\ref{sec:behavioral} first tests the predicted behavioral distinctions. \S\ref{sec:causal-flow} then provides our main analysis: whether the corresponding variables align with causally effective subspaces and whether the matching and relative polarity interact, without assuming a specific low-level operation or context-invariant neural representations.


\section{Can Models Even Solve Negation?}
\label{sec:behavioral}

We first assess whether LLMs can solve the controlled NLI setting and exhibit the behavior predicted by the Polarity Composition Hypothesis (\S\ref{sec:polarity-hypothesis}). In this section, this assessment proceeds in three stages: we isolate polarity comparison under a fixed event match (\S\ref{sec:beh-polarity-comp}), test if event matching gates this comparison (\S\ref{sec:beh-matching}), and evaluate assumption selection in the full multi-event setting (\S\ref{sec:beh-select}).

We evaluate Qwen3-8B~\citep{qwen3technicalreport} and Phi-4 Mini Instruct~\citep{microsoft2025phi4minitechnicalreportcompact}, abbreviated as Phi-4 Mini in tables. We evaluated seven open-weight models of similar size on the three tests below (Appendix~\ref{apx:additional-models}, Table~\ref{tab:additional-model-screening}). Three models exceed 85\% accuracy on all tests: Qwen3-8B, Phi-4 Mini Instruct, and Gemma 3 12B. We run the full layer-wise analysis on the first two, which is the limit of our compute budget. The other four stay below 60\% on mismatched events, so the task is not trivial at this scale. All prompts follow Figure~\ref{fig:prompt-template}; the first two experiments use one assumption event, and the third uses three. We extract next-token logits of the single-token labels ``$\mathrm{T}$,'' ``$\mathrm{F}$,'' and ``$\mathrm{U}$'' and take their argmax; no free-form text is generated.

We use $R=\operatorname{logit}(\mathrm{T})-\operatorname{logit}(\mathrm{F})$ to measure the direction of the $\mathrm{T}/\mathrm{F}$ preference: $R>0$ favors $\mathrm{T}$, whereas $R<0$ favors $\mathrm{F}$. We use $G_U=\operatorname{logit}(\mathrm{U})- \max\{\operatorname{logit}(\mathrm{T}), \operatorname{logit}(\mathrm{F})\}$ to measure preference for uncertainty: $G_U>0$ favors $\mathrm{U}$, whereas $G_U<0$ favors a polarized $\mathrm{T}/\mathrm{F}$ decision. 
Every experiment starts from 1,000 SVO events sampled from the lexical inventory in Appendix~\ref{apx:data_construct}; prompt counts differ because each experiment expands these events by a different design (Appendix~\ref{apx:behavioral}). Results are averaged over three independently generated sets. Negative polarity uses \textit{did not}; \S\ref{sec:robustness} tests alternative forms.


\subsection{Polarity Composition}
\label{sec:beh-polarity-comp}
We first isolate polarity comparison by holding event content fixed. We sample 1,000 matching SVO events ($m=1$) with four polarity combinations, yielding 4,000 prompts. Here $A^{+}/A^{-}$ and $C^{+}/C^{-}$ denote affirmative/negative assumptions and claims. The example in \S\ref{sec:polarity-hypothesis} illustrates $A^{-}C^{+}$. Since event content matches, equal-polarity conditions ($A^{+}C^{+}$ and $A^{-}C^{-}$) should favor $\mathrm{T}$ ($R>0$), whereas opposite-polarity conditions ($A^{+}C^{-}$ and $A^{-}C^{+}$) should favor $\mathrm{F}$ ($R<0$). 

\begin{table}[t]
\centering
\caption{\textbf{Polarity composition.} Accuracy (\%)/mean $R$. Positive and negative $R$ favor $\mathrm{T}$ and $\mathrm{F}$, respectively.}
\label{tab:behavioral-polarity}
\small
\setlength{\tabcolsep}{1.3pt}
\begin{tabular}{@{}lrrrr@{}}
\toprule
Model & $A^{+}C^{+}$ & $A^{-}C^{-}$ & $A^{+}C^{-}$ & $A^{-}C^{+}$  \\
\midrule
Qwen3-8B
& 99.9/8.20
& 100.0/7.96
& 100.0/$-$4.93
& 100.0/$-$5.17
 \\
Phi-4 Mini
& 99.9/5.45
& 84.6/2.90 
& 98.1/$-$4.56
& 90.5/$-$5.82
\\
\bottomrule
\end{tabular}
\end{table}

Table~\ref{tab:behavioral-polarity} shows that mean $R$ has the expected sign in all four polarity conditions for both models. Qwen3-8B achieves near-perfect accuracy throughout. Phi-4 Mini preserves the expected T/F ordering, but its accuracy decreases in negative cases. 

\subsection{Matching as a Polarity Gate}
\label{sec:beh-matching}
We next test whether event matching determines whether polarity affects the output. We compare full SVO agreement with partial and complete mismatch across the four polarity combinations (20,000 prompts). If matching gates polarity-based inference, full matches should yield $G_U <0$, whereas mismatches should yield $G_U>0$.

Figure~\ref{fig:behavioral-gate} shows a consistent sign reversal. Under full matching, mean $G_U$ is $-3.35$ for Qwen3-8B and $-1.90$ for Phi-4 Mini Instruct. Across mismatch conditions, both models show positive $G_U$. Therefore, both models favor a polarized T/F decision for matching events and redirect mismatched events toward U.
\begin{table}[t]
\centering
\caption{\textbf{Relevant-assumption selection.}
Target and Distr.\ : mean absolute change in $R$ after modifying target/distractor polarity. Stable: percentage of distractor modifications that preserve the predicted label. Ratio $=$ Target/Distr.}
\label{tab:behavioral-selection}

\small
\setlength{\tabcolsep}{1.5pt}
\begin{tabular}{@{}lrrrrr@{}}
\toprule
Model & Acc. (\%) & Target & Distr. & Stable (\%) & Ratio \\
\midrule
Qwen3-8B
& 98.94 & 14.29 & 0.39 & 98.88 & 37.0$\times$ \\
Phi-4 Mini
& 89.37 & 10.19 & 0.66 & 92.06 & 15.3$\times$ \\
\bottomrule
\end{tabular}
\end{table}
\subsection{Selecting the Relevant Assumption}
\label{sec:beh-select}
\begin{table*}
\centering
\caption{\textbf{Variable-identifying controls.}
``Reverse'' and ``preserve'' refer to the base $\mathrm{T}/\mathrm{F}$ label under an open gate. The $\rho_i$ controls identify the relation
within a fixed matching state; cross-state transfer is evaluated separately in \S\ref{sec:rho-context}.}
\label{tab:identity-controls}
\small
\setlength{\tabcolsep}{4pt}
\begin{tabular}{
    @{}lp{0.33\textwidth}p{0.24\textwidth}p{0.27\textwidth}@{}
}
\toprule
Target & Source--base contrast & Expected target & Excludes \\
\midrule

$p_c$
& Change $p_c$; flip both polarities; flip only $p_i$
& Reverse; reverse; preserve
& $p_i$, $\rho_i$, and source-label copying \\

$p_i$
& Change $p_i$; flip both polarities; flip only $p_c$;
  use a closed-gate source
& Reverse; reverse; preserve; follow
  $g(p_i^{(s)}p_c^{(b)})$
& $p_c$, $\rho_i$, gated relation, and source-label copying \\

$\rho_i$
& Within one matching state: flip either polarity; flip both
& Reverse; preserve
& Either raw polarity \\

$m_i$
& Match $\rightarrow$ mismatch; mismatch $\rightarrow$ match;
  anti-copy sources
& $\mathrm{U}$; $g(\rho_i^{(b)})$; base-context label
& Source polarity, content, and direct label copying \\

\bottomrule
\end{tabular}
\end{table*}
Finally, we test whether models use the polarity of the matching assumption rather than those of distractors. Each prompt contains one matching assumption (the target) and two distractors, with target position and polarity balanced (18,000 prompts). We add or remove negation from either the target or a distractor and measure the resulting change in $R$. If the model selects the matching assumption, $R$ should change substantially when the target polarity is modified but remain nearly unchanged when a distractor is modified.



In Table~\ref{tab:behavioral-selection}, for both models, adding or removing negation from the target moves $R$ in the expected direction.  Target modifications have substantially larger effects than distractor modifications, with target-to-distractor ratios of 37.0$\times$ for Qwen3-8B and 15.3$\times$ for Phi-4 Mini. 
Overall, both models base their decisions on the matching assumption.

Together, the experiments show that both models exhibit the behavioral distinctions captured by the Polarity Composition Hypothesis: $R$ follows relative polarity for matching events, $G_U$ separates matched from mismatched events, and multi-event decisions are substantially more sensitive to the matching assumption than to distractors. These results do not identify the underlying mechanism. \S\ref{sec:causal-flow} therefore tests whether the corresponding variables align with causally effective representations.

\section{Causality of Polarity Composition}
\label{sec:causal-flow}
\begin{figure}[t]
    \centering
    \includegraphics[
        width=0.95\columnwidth,
        trim=0.2cm 9cm 2cm 0.5cm,
        clip
    ]{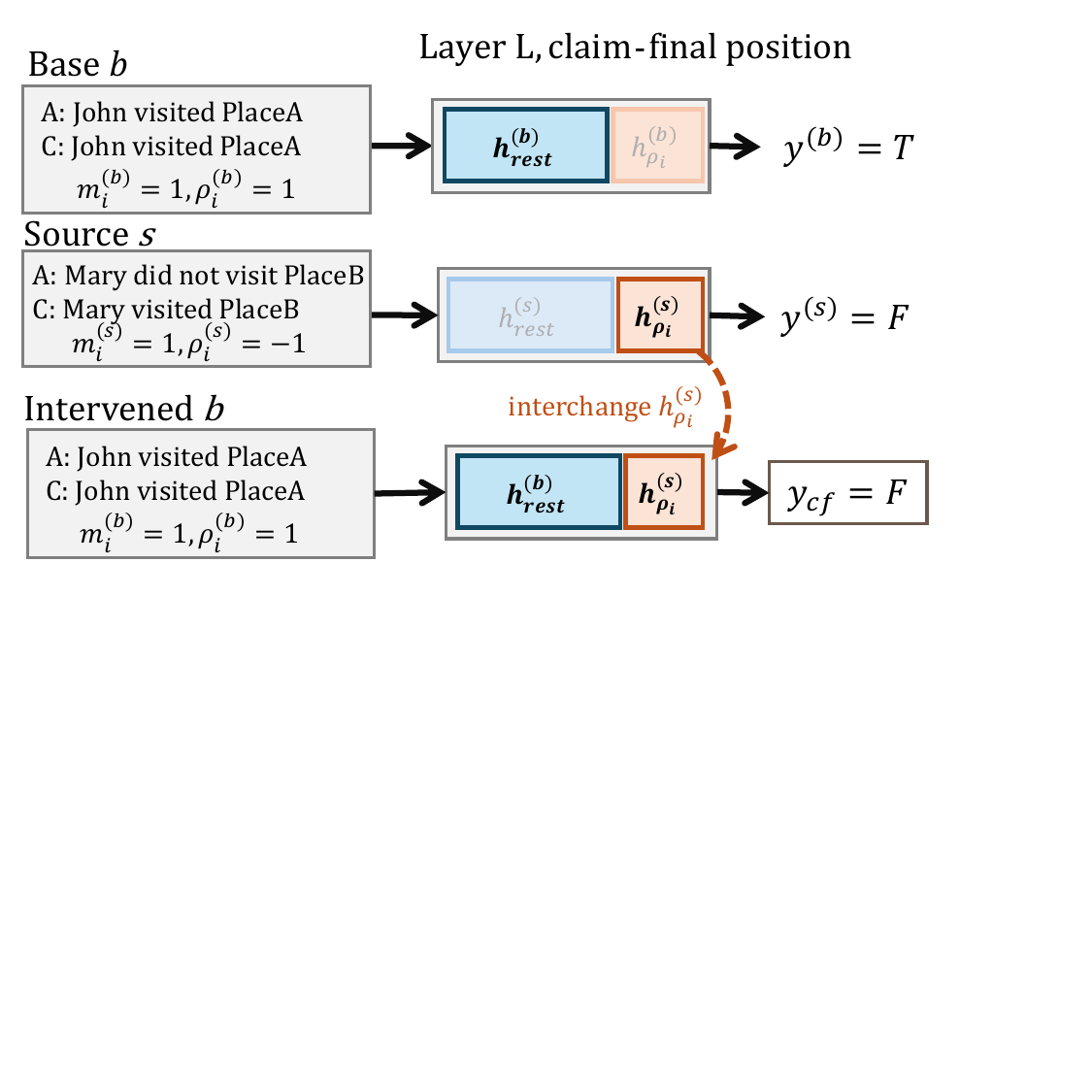}

    \caption{\textbf{DAS intervention on relative polarity.} For $v=\rho_i$, DAS interchanges $h_{\rho_i}^{(b)}$ with $h_{\rho_i}^{(s)}$ while retaining $h_{\mathrm{rest}}^{(b)}$. The unchanged base input therefore yields counterfactual $y_{cf}=\mathrm{F}$ instead of $\mathrm{T}$.}
    \label{fig:das}
\end{figure}
The behavioral results (\S \ref{sec:behavioral}) are consistent with Polarity Composition Hypothesis in Equation~\ref{eq:causal-flow}, which validates the positive correlation between the hypothesis and actual model behavior. We now test whether four variables defined in \S\ref{sec:polarity-hypothesis}: assumption polarity $p_i$, claim polarity $p_c$, relative polarity $\rho_i$, and event matching $m_i$ align with causally effective subspaces. We first identify and localize individual variables (\S\ref{sec:causal-layer-wise}), test whether matching and relation subspace jointly control the output (\S\ref{sec:joint-composition}), and finally examine whether relation representation transfers across matching contexts (\S\ref{sec:rho-context}).

\subsection{Experimental Setting}
\label{sec:das-setting}

\paragraph{Distributed Alignment Search} 
We use DAS \citep{geiger-etal-2024-finding} to align each high-level variable with a subspace of the model's internal representation.  Each intervention pairs a base prompt $b$, which provides the surrounding context, with a source prompt $s$, which provides the intended value of the target variable.
At a selected layer and token position, let $h^{(b)},h^{(s)}\in\mathbb{R}^{d}$ denote the hidden states produced by the unmodified model for the base and source prompts, respectively. Figure~\ref{fig:das} illustrates the intervention for $\rho_i$. 

For target variable $v$, let $B_v\in\mathbb{R}^{d\times k}$ be an
orthonormal basis for a $k$-dimensional learned subspace. DAS
intervenes on the base representation as
\begin{equation}
    \widetilde{h}^{(b)}
    =
    h^{(b)}
    +
    B_vB_v^\top
    \left(h^{(s)}-h^{(b)}\right),
    \label{eq:das-intervention}
\end{equation}
With $h_v^{(x)}=B_vB_v^\top h^{(x)}$ and
$h_{\mathrm{rest}}^{(x)}=h^{(x)}-h_v^{(x)}$, this update becomes
$\widetilde h^{(b)}=h_{\mathrm{rest}}^{(b)}+h_v^{(s)}$: the source
component replaces the base component while the base remainder is
retained. The language model remains frozen; only $B_v$ is optimized.

We optimize each subspace for one epoch using AdamW with a learning rate of $2\times10^{-3}$ and rank $k=64$ for both models.\footnote{Robustness to alternative subspace ranks is reported in Appendix~\ref{apx:rank-robustness}.}
We learn a separate subspace at every second layer and evaluate at two token positions: the final token of the claim (claim-final) and the token after \texttt{Answer:} (answer token).\footnote{For $p_i$, we additionally evaluate the final token of the corresponding assumption event, where the causal attention mask prevents access to the subsequent claim polarity $p_c$. Results are in Appendix~\ref{apx:pi_subspace}.} Layer indices are normalized by model depth.

\paragraph{Data} For each variable, we sample 1,000 SVO events from the inventory in Appendix~\ref{apx:data_construct} and split them by complete tuple into 700/150/150 training, validation, and test events. All prompts from an event remain in the same split, and held-out events are excluded as training distractors or mismatches. Training uses variable-specific controls; validation and test controls are balanced.\footnote{Dataset sizes and sampling proportions are provided in Appendix~\ref{apx:das-controls}.}

\paragraph{Intervention Design} For a target variable $v\in\{p_c,p_i,\rho_i,m_i\}$, superscripts $(b)$ and $(s)$ denote values from the base and source prompts. The source supplies the value of $v$, while the other arguments retain their base values:

\begin{equation}
y_{\mathrm{cf}}(v)=
\begin{cases}
g(m_i^{(b)}p_i^{(b)}p_c^{(s)}), & v=p_c,\\
g(m_i^{(b)}p_i^{(s)}p_c^{(b)}), & v=p_i,\\
g(m_i^{(b)}\rho_i^{(s)}),       & v=\rho_i,\\
g(m_i^{(s)}\rho_i^{(b)}),       & v=m_i.
\end{cases}
\label{eq:atomic-counterfactuals}
\end{equation}


A single successful contrast does not uniquely identify the intervened variable: a subspace may instead encode a raw polarity, the source label, or another correlated feature. We therefore evaluate each candidate subspace using the identifying controls summarized in Table~\ref{tab:identity-controls}.

The polarity datasets additionally include closed-base controls, which must remain $\mathrm{U}$ even when the source carries a polarized label; $p_i$ also includes a non-matching-assumption locality control. For $m_i$, anti-copy conditions make the source label disagree with the counterfactual label required by the base context. As in \S\ref{sec:polarity-hypothesis}, the $m_i$ counterfactual assumes at most one matching assumption.

\paragraph{Evaluation} Interchange intervention accuracy (IIA) is the proportion of intervened outputs matching the counterfactual label in Equation~\ref{eq:atomic-counterfactuals} \citep{geiger2025causal}. For each layer-position-control, IIA is averaged over three DAS seeds. Confirmatory IIA is the minimum across the identifying controls in Table~\ref{tab:identity-controls}, preventing a high average from hiding a failed contrast. For $\rho_i$, it uses only within-state controls; cross-state transfer is audited in \S\ref{sec:rho-context}.

\paragraph{Necessity} To test whether the model depends on a learned subspace, we resample its coordinates from randomly selected test examples without conditioning on the source value. We compare this intervention with resampling an equal-rank random subspace. For target variable $v$, we define 

\begingroup
\small
\setlength{\abovedisplayskip}{1pt}
\setlength{\belowdisplayskip}{3pt}
\begin{equation}
\operatorname{NEx}(v)
=
\frac{1}{|\mathcal{C}_v|}
\sum_{c\in\mathcal{C}_v}
\left(
\operatorname{Acc}_{\mathrm{random},c}
-
\operatorname{Acc}_{\mathrm{DAS},c}
\right).
\label{eq:nex}
\end{equation}
\endgroup
where $\mathcal{C}_v$ contains the relevant identifying controls. Positive NEx indicates that disrupting the learned subspace is more harmful than disrupting a rank-matched random subspace.\footnote{Value-conditioned results are reported in Appendix~\ref{apx:polarity-resampling}.}

\begin{figure*}[ht]
    \centering
    \includegraphics[width=0.92\textwidth]{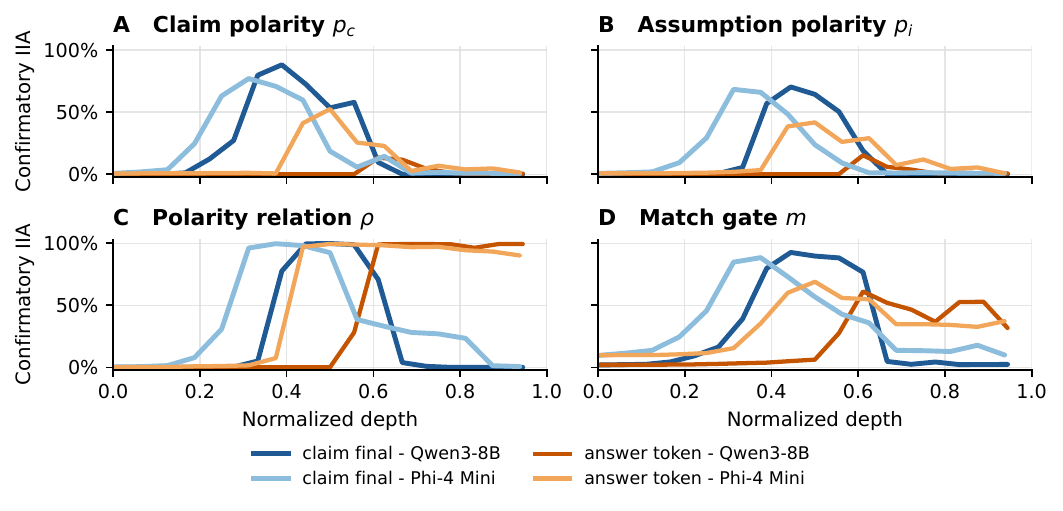}
    \caption{\textbf{Layer-wise causal identification of polarity variables.}
    Confirmatory Interchange Intervention Accuracy (IIA) across normalized model depth at the claim-final and answer-token positions. High confirmatory IIA indicates a causally effective subspace that consistently identifies the target variable.}
    \label{fig:causal-flow}
\end{figure*}

\subsection{Layer-wise Causal Representations}
\label{sec:causal-layer-wise}

\begin{table}[t]
\centering
\caption{\textbf{Selected causal representations.}
IIA is in \% and NEx in pp; $\rho_i$ IIA uses within-state controls. $^\dagger$ Weak raw-polarity identity, so NEx is not attributed specifically to raw polarity.}
\label{tab:causal-peaks}
\small
\setlength{\tabcolsep}{1.5pt}
\begin{tabular}{@{}llrrr@{\hspace{3pt}}rrr@{}}
\toprule
& & \multicolumn{3}{c}{Claim}
& \multicolumn{3}{c}{Answer} \\
\cmidrule(lr){3-5}\cmidrule(lr){6-8}
Model & Var.
& L & IIA & NEx
& L & IIA & NEx \\
\midrule
Phi-4 Mini & $p_c$
& 10 & 77.0 & 35.7
& 16 & 52.4 & 34.7 \\
Qwen3-8B & $p_c$
& 14 & 88.1 & 44.1
& 22 & 13.3 & 37.5$^\dagger$ \\
\addlinespace[1pt]

Phi-4 Mini & $p_i$
& 10 & 68.3 & 45.5
& 16 & 41.7 & 44.0$^\dagger$ \\
Qwen3-8B & $p_i$
& 16 & 70.2 & 49.8
& 22 & 15.3 & 52.8$^\dagger$ \\
\addlinespace[1pt]

Phi-4 Mini & $\rho_i$
& 12 & 99.8 & 50.2
& 16 & 99.4 & 54.0 \\
Qwen3-8B & $\rho_i$
& 18 & 99.9 & 59.6
& 24 & 99.3 & 64.4 \\
\addlinespace[1pt]

Phi-4 Mini & $m_i$
& 12 & 88.4 & 42.8
& 16 & 68.9 & 45.3 \\
Qwen3-8B & $m_i$
& 16 & 90.7 & 49.6
& 22 & 62.9 & 48.7 \\
\bottomrule
\end{tabular}
\end{table}

\paragraph{Raw Polarities $p_i,p_c$}
As shown in Figure~\ref{fig:causal-flow}, both models contain identifiable $p_c$ at claim-final. Confirmatory IIA reaches 77.0\% and 88.1\% for $p_c$, and 68.3\% and 70.2\% for $p_i$, in Phi-4 Mini Instruct and Qwen3-8B respectively. These scores decrease substantially at the answer token, suggesting that raw-polarity representations are localized near where the corresponding information is initially processed, rather than uniformly. This localization is examined further across negation forms in \S\ref{sec:robust-negation}. Table~\ref{tab:causal-peaks} also reports NEx at the selected peak layers. At claim final, raw-polarity NEx ranges from 35.7 to 49.8 pp, indicating substantially greater disruption from resampling the learned subspaces than from resampling equal-rank random subspaces. 

\paragraph{Relative Polarity $\rho_i$}
Under the within-state identity controls, the learned $\rho_i$ subspace remains near the ceiling from the claim final to the answer token in both models. Its NEx ranges from 50.2 to 64.4 pp, indicating substantially greater disruption than an equal-rank random subspace. These results identify a causal \(\rho_i\)-aligned comparison/decision state within matching contexts, where it is indistinguishable from an abstract T/F decision. 
\paragraph{Event Matching $m_i$}
At the claim-final position, $m_i$ is identified strongly(88.4-90.7\%), indicating a causally effective gate rather than being a source-label or source-polarity copying. At the answer token, IIA reduces to 62-68\%, while NEx remains high. This pattern suggests that $m_i$ is explicit after assumption-claim comparison but becomes integrated with $\rho_i$ in the downstream: its separate identity weakens at the answer while its causal influence remains.

\subsection{Causal Interaction of $m_i$ and $\rho_i$}
\label{sec:joint-composition}
Identifying $m_i$ and $\rho_i$ separately does not establish that the model composes them. We perform joint interventions at the claim-final position using two independently selected sources:
\begin{equation}
    y_{\mathrm{cf}}
    =
    g\!\left(m_i^{(s_m)}\rho_i^{(s_\rho)}\right).
    \label{eq:joint-gate}
\end{equation}
This strict gate-opening condition begins with $m_i^{(b)}=0$. One source provides an open gate, while the $\rho_i$ source is selected from $m_i^{(s_\rho)}=0$ and therefore has the natural label $\mathrm{U}$. Recovering the target $\mathrm{T}/\mathrm{F}$ label thus requires two subspaces to interact. Layers are selected by validation confirmatory IIA. This selects L12 for both variables\footnote{patch using a minimum-norm update to avoid overlap} in Phi-4 Mini Instruct, and L16/L18 for $m_i$/$\rho_i$ in Qwen3-8B\footnote{patch applying within same forward pass}. 
\begin{table}[t]
\centering
\caption{\textbf{Strict gate-opening composition.} Best single writes only $m_i$ or $\rho_i$ subspace; Joint writes both.}
\label{tab:joint-composition}
\setlength{\tabcolsep}{4pt}
\begin{tabular}{@{}llrr@{}}
\toprule
Model & Target & Best single & Joint \\
\midrule
Phi-4 Mini& $\mathrm{F}$ & 0.5 & 62.8 \\
Phi-4 Mini & $\mathrm{T}$ & 12.5 & 92.3 \\
Qwen3-8B            & $\mathrm{F}$ & 0.7 & 92.3 \\
Qwen3-8B            & $\mathrm{T}$ & 2.8 & 98.5 \\
\bottomrule
\end{tabular}
\end{table}

Table~\ref{tab:joint-composition} shows that the joint intervention substantially outperforms either intervention alone, supporting an interaction between $m_i$ and $\rho_i$. However, because $\rho_i$ is copied from a source with $m_i=0$, we do not know whether it retains the same meaning after the gate is opened. Thus, the experiment does not establish a context-invariant $\rho_i$ representation or literal multiplication of $m_i$ and $\rho_i$.

\subsection{Transfer of $\rho_i$ depends on $m_i$}
\label{sec:rho-context}

We test whether the same value of $\rho_i$ transfers consistently across matching states. All bases have $m_i^{(b)}=1$. We patch only the learned $\rho_i$ subspace and cross whether the source has the same or opposite $m_i$ and $\rho_i$ as the base. Thus, opposite-$m_i$/same-$\rho_i$ means $m_i^{(s)}=0$ and $\rho_i^{(s)}=\rho_i^{(b)}$, while the base matching state remains unchanged. We evaluate the $\rho_i$ representations reported in Table~\ref{tab:causal-peaks}: L12/L16 for Phi-4 Mini Instruct and L18/L24 for Qwen3-8B at the claim-final/answer-token positions.

In Table \ref{tab:rho-context}, when source and base share the same matching state, same-$\rho_i$ patches largely preserve the base label, while opposite-$\rho_i$ patches reverse more than 99\% of cases. With cross-state sources, same-$\rho_i$ preservation falls to 0.1-13.5\%, although opposite-$\rho_i$ transfer remains high. Success on the opposite-$\rho_i$ condition alone is insufficient to establish context-invariant transfer, which requires both relation values to be preserved consistently.

These results show that the learned $\rho_i$ subspace transfers reliably within a matching state but not across matching states. This asymmetry suggests that event matching conditions either the construction or downstream interpretation of relation information. Because the cross-state same-$\rho_i$ condition was not used during DAS optimization, its failure reveals a limitation of the learned alignment without ruling out a more context-invariant subspace.

\begin{table}[t]
\centering
\caption{\textbf{Factorial audit of $\rho_i$ transfer (\%).} Entries report expected-effect accuracy: base-label preservation for same-$\rho_i$ and reversal for opposite-$\rho_i$. C/A denote claim-final and answer-token positions.}
\label{tab:rho-context}
\small
\setlength{\tabcolsep}{1.5pt}
\begin{tabular}{@{}llrrrr@{}}
\toprule
& & \multicolumn{2}{c}{Phi-4 Mini}
& \multicolumn{2}{c}{Qwen3-8B} \\
\cmidrule(lr){3-4}\cmidrule(lr){5-6}
$m_s:m_b$ & $\rho_s:\rho_b$
& L12 C & L16 A & L18 C & L24 A \\
\midrule
Same & Same
& 85.5 & 85.5 & 98.4 & 98.3 \\
Same & Opp.
& 99.6 & 99.6 & 99.9 & 99.9 \\
Opp. & Same
& 13.5 & 1.6 & 11.0 & 0.1 \\
Opp. & Opp.
& 99.6 & 99.7 & 99.9 & 99.9 \\
\bottomrule
\end{tabular}
\end{table}

\section{Robustness}
\label{sec:robustness}
We evaluate whether the behavioral and causal structure identified in the controlled setting generalizes across negation forms and to naturalistic NLI examples. All causal evaluations reuse the DAS subspaces identified in \S\ref{sec:causal-flow}. No rotation is retrained on the robustness data. Answer-token and variable-specific results are provided in Appendix~\ref{apx:robustness}.

\subsection{Robustness to Negation Form}
\label{sec:robust-negation}
Beyond the canonical \textit{did not}, we test whether the same computation generalizes to \textit{didn't}, \textit{never}, \textit{did not ever}, \textit{failed to}, and \textit{it is not the case that}. Within the open-gate setting ($m_i=1$) used here, the frozen $\rho_i$ subspace transfers strongly across forms: confirmatory IIA is at least 96.8\% within form and at least 98.3\% from \textit{did not} to an alternative form. Its within-state transfer is more stable than Phi-4 Mini Instruct's end-to-end behavior (Appendix~\ref{apx:robustness}). Because the subspace was learned only on \textit{did not}, this transfer suggests that the $\rho_i$-aligned state does not depend on the surface form of negation, including \textit{failed to}, which contains no negation word.

\subsection{Naturalistic Transfer to MNLI}
\label{sec:robust-mnli}
MNLI \citep{williams-etal-2018-broad} is a broad-coverage NLI corpus of naturally occurring sentence pairs; we use it to test whether subspaces learned on synthetic data transfer to natural text. Following \citet{hossain-etal-2020-analysis}, we recover 155 complete MNLI polarity squares whose labels follow $(\mathrm{T},\mathrm{F},\mathrm{F},\mathrm{T})$, yielding 620 examples. We additionally use 101 all-$\mathrm{U}$ squares as a closed-gate comparison. MNLI may have appeared in the models' pretraining data, but this is not a barrier to our investigation: our analysis asks which internal variables causally control a prediction, not whether it reflects generalization or memorization.

Table~\ref{tab:robust-mnli-main} summarizes behavioral performance and causal transfer. We use the answer token as primary causal position, where assumption-claim information has been integrated into the output decision. For same-$\rho_i$ interventions, the counterfactual target preserves the base relation label; for opposite-$\rho_i$ interventions, it reverses the label. Exact IIA requires the target label to be T/F/U argmax, whereas T/F IIA evaluates the direction of the T/F preference.
\begin{table}[t]
\centering
\caption{\textbf{MNLI robustness.}
Behavioral accuracy and answer-token IIA (\%).}
\label{tab:robust-mnli-main}
\small
\setlength{\tabcolsep}{3.5pt}
\begin{tabular}{@{}lrr@{}}
\toprule
Metric & Phi-4 Mini & Qwen3-8B \\
\midrule
Row accuracy                         & 54.8 & 80.2 \\
Complete-square accuracy             & 18.1 & 54.8 \\
\midrule
Same-$\rho_i$ exact IIA              & 60.4 & 81.6 \\
Same-$\rho_i$ T/F IIA                & 84.2 & 94.6 \\
Opposite-$\rho_i$ exact IIA          & 65.3 & 83.2 \\
Opposite-$\rho_i$ T/F IIA            & 84.6 & 91.4 \\
$m_i$: polarized $\rightarrow\mathrm{U}$ IIA
                                      & 88.9 & 77.9 \\
$m_i$: $\mathrm{U}\rightarrow$ polarized IIA
                                      & 65.7 & 63.9 \\
\bottomrule
\end{tabular}
\end{table}

At the answer token, same-$\rho_i$ interventions preserve the T/F direction in 84.2-94.6\% of cases, while opposite-$\rho_i$ interventions reverse it in 84.6-91.4\%. Matching interventions move predictions between polarized and $\mathrm{U}$ outputs at 63.9-88.9\% IIA, whereas $\rho_i$ interventions with the same sources do so in only 5.2-14.0\% of cases. Thus, $\rho_i$ primarily controls the T/F relation, whereas $m_i$ controls whether the output is polarized or unknown.

\section{Discussion}
\label{sec:discussion}

Our results suggest a staged computation. Raw polarities are most identifiable near where they are read, whereas the matching state and relative polarity remain causally effective later. Joint intervention shows that the learned matching and relation subspace interact at the claim-final position. Here, matching refers to exact SVO identity, tested with balanced partial- and no-overlap controls. This interaction supports the proposed gating structure, but does not identify its exact neural operation.

Our conclusions about $\rho_i$ are strongest when the gate is open, and the assumption event matches the claim. In this setting, the relative-polarity alignment is identifiable, causally effective, and interacts with the matching subspace. When the gate is closed, $m_i$ already determines the output as $\mathrm{U}$ regardless of $\rho_i$. The model, therefore, need not use a relative-polarity value, and its output does not reveal whether or how such a value is represented. Consistently, the learned $\rho_i$ alignment does not transfer reliably between matching states. Thus, $\rho_i$ may be formed only for matching events, or represented or interpreted differently when no event matches; our experiments do not distinguish between these possibilities.

Across negation forms and the MNLI subset, the frozen $\rho_i$ alignment transfers more reliably than the models' end-to-end behavior. On MNLI, opening the matching gate ($\mathrm{U}\rightarrow$ polarized) is weaker than closing it, suggesting that event matching may be an important bottleneck in naturalistic transfer. 

\section{Conclusion}

We propose and causally test Polarity Composition Hypothesis to explain LLMs internal mechanism when solving NLI under negation, in which event matching gates relative polarity. DAS identifies event matching and relative polarity subspaces whose joint intervention recovers strict gate-opening targets more reliably than either alone. The relative polarity subspace persists toward the answer and transfers across negation constructions, and partially to MNLI, but its behavior is not reliably interchangeable across matching states. These results support gated polarity agreement without establishing a context-invariant relation code or a complete natural-logic implementation.
\section*{Limitations}
Our controlled grammar covers simple English SVO events and assumes at most one matching assumption. With a quantified subject, as in \textit{No one visited PlaceA} against \textit{John visited PlaceA}, the contents differ and the hypothesis returns \textsc{Unknown}, although the correct label is \textsc{False}; such quantified negation, and negative-concord languages, lie outside this design. DAS identifies causally effective low-rank directions, not unique intrinsic variables, and the cross-state results show that their intervention semantics can depend on context. The MNLI evaluation uses a curated subset, and we study only two models.

Under the single-match design, $\rho_i$ is aligned with the T/F direction when the gate is open, while $m_i$ is aligned with the polarized-versus-U distinction. Our controls distinguish these channels from raw polarity and direct source-label copying, but do not uniquely exclude all downstream decision-aligned representations.
\section*{AI Use Statement}
We used LLMs to assist with manuscript editing, including grammar, spelling, and rephrasing, and with code development. We manually reviewed all generated text and code for correctness and security. All scientific claims, interpretations, and remaining errors are the authors' responsibility.

\bibliography{custom}

\appendix

\section{Hardware}
All experiments were conducted on NVIDIA RTX A6000 GPUs with 48\,GB of memory in the cloud environment. We used three random seeds(0,1,2) to assess the variability.
\section{Dataset Construction}
\label{apx:data_construct}

\subsection{Lexical Inventory}

Each atomic event is an SVO tuple $e=(S,V,O)$. Subjects are sampled from 50 people's names. Table~\ref{tab:lexical-inventory} reports the complete lexical inventory.

\begin{table*}[t]
\centering
\caption{\textbf{Lexical inventory.} The six additional verbs are
used only for the $m_i$ dataset.}
\label{tab:lexical-inventory}
\small
\setlength{\tabcolsep}{5pt}
\begin{tabular}{@{}p{0.19\textwidth}p{0.74\textwidth}@{}}
\toprule
Component & Inventory \\
\midrule
Subjects
&
Jack, Luke, Noah, Liam, Owen, Ryan, Adam, Evan, Leo, Ian,
Eric, John, Paul, Dean, Martin, Cody, Brian, Mike, Thomas,
Chris, Mark, Tyler, Jake, Kyle, Zane, Emma, Ella, Ava, Mia,
Lily, Zoe, Chloe, Grace, Lucy, Anna, Clara, Nora, Ruby, Ivy,
Alice, Eva, Jane, Rose, Lisa, Sara, Kate, Jean, Claire,
Amber, Sophie
\\

Common place verbs
&
\texttt{visit/visited}, \texttt{reach/reached},
\texttt{explore/explored}, \texttt{enter/entered},
\texttt{like/liked}
\\

Common object verbs
&
\texttt{open/opened}, \texttt{create/created},
\texttt{clean/cleaned}, \texttt{collect/collected},
\texttt{close/closed}
\\

Additional $m_i$ place verbs
&
\texttt{mention/mentioned}, \texttt{recommend/recommended},
\texttt{describe/described}
\\

Additional $m_i$ object verbs
&
\texttt{inspect/inspected}, \texttt{paint/painted},
\texttt{move/moved}
\\

Arguments
&
\texttt{PlaceA, PlaceB, PlaceC}; \texttt{ObjectA, ObjectB, ObjectC}
\\
\bottomrule
\end{tabular}
\end{table*}

The $p_i$, $p_c$, and $\rho_i$ datasets use the ten common verbs, giving $50\times10\times3=1{,}500$ possible SVO events. The $m_i$ dataset uses all 16 verbs, giving $50\times16\times3=2{,}400$ possible events. Each dataset samples 1,000 distinct base events without replacement.

For $m_i$, verb mismatches preserve the argument type and are drawn between fixed, disjoint verb pools. The place-compatible pools are $\{\texttt{visit},\texttt{reach},\texttt{explore},\texttt{enter}\}$ and $\{\texttt{like},\texttt{mention},\texttt{recommend}, \texttt{describe}\}$; the object-compatible pools contain the five common and three additional object verbs, respectively.

\subsection{Event and Mismatch Construction}

In the canonical controlled datasets, affirmative and negative
events use
\begin{align}
    e^{+} &: \quad \texttt{S V\textsubscript{past} O.},\\
    e^{-} &: \quad \texttt{S did not V\textsubscript{base} O.}
\end{align}
For example:
\begin{align}
    e^{+} &= \texttt{Jack visited PlaceA.},\\
    e^{-} &= \texttt{Jack did not visit PlaceA.}
\end{align}

For the $m_i$ DAS dataset, the four mismatch types relative to a
claim event $(S,V,O)$ are
\begin{align}
    \text{subject mismatch} &: (S',V,O),\\
    \text{verb mismatch} &: (S,V',O),\\
    \text{object mismatch} &: (S,V,O'),\\
    \text{no overlap} &: (S',V',O'),
\end{align}
where every primed component differs from its claim counterpart.
Clean distractors and no-overlap replacements differ from the claim
in all three SVO components and are sampled from the ten-verb common
inventory.

Base events are split by complete SVO tuple into 700/150/150
training, validation, and test events. Training rows exclude
validation and test base events when sampling auxiliary events, and
validation rows exclude test base events. Rows for which this
restriction must be relaxed are flagged during generation and
omitted from DAS training and evaluation.

For $m_i$, training retains all remaining mismatch realizations.
Validation and test retain one mismatch realization per polarity
and control cell, while preserving the additional
\texttt{label\_copy\_trap\_same\_m1} control. This prevents aggregate
IIA from being dominated by the four mismatch types.

\subsection{Prompt Template}

Each DAS input contains three assumption events and one hypothesis
event, displayed as the \texttt{Claim}. The designated target
position is balanced across the three assumption positions. It contains
an exact SVO match when $m_i=1$ and a mismatch event when $m_i=0$.
Base and source inputs use the same template:

\begin{small}
\begin{verbatim}
Assumption:
1. <assumption event 1>
2. <assumption event 2>
3. <assumption event 3>

Claim: <claim event>

Choose exactly one:
T = must be true
F = must be false
U = cannot be determined

Answer:
\end{verbatim}
\end{small}

Predictions are obtained from the next-token logits of \texttt{T}, \texttt{F}, and \texttt{U} after \texttt{Answer:}. Single-assumption behavioral experiments use the same template with one assumption.

\section{Behavioral Experiment Details}
\label{apx:behavioral}
\subsection{Polarity Composition}

Each generated set contains 1,000 matching SVO events crossed with the four assumption-hypothesis polarity combinations, yielding $1{,}000\times4=4{,}000$ prompts. Results are averaged over three independently generated sets.

\subsection{Matching Gate}
We cross 1,000 events with four polarity combinations and five content conditions: SVO, SV, VO, subject-only, and no overlap, yielding $1{,}000\times4\times5=20{,}000$ prompts. 
\subsection{Relevant-Assumption Selection}

For each of 1,000 base events, the claim is affirmative and each prompt contains one SVO-matching assumption and two distractors that share no subject, verb, or object with the claim. We cross three target positions, two target polarities, and three distractor configurations: unchanged, first distractor flipped, and second distractor flipped. This yields $1{,}000\times3\times2\times3=18{,}000$ prompts.

Pairing opposite target polarities while holding everything else fixed gives 9,000 target-flip pairs. Comparing each distractor flip with its unchanged counterpart gives 12,000 distractor-flip pairs. We report mean $|\Delta R|$ for both pair types, target-flip directional accuracy, and distractor-flip prediction invariance. The target-to-distractor ratio is computed from unrounded mean
effects.

\subsection{Zero-Shot Behavioral Screening}
\label{apx:additional-models}

In Table \ref{tab:additional-model-screening}, we test whether similarly sized language models can reliably solve our controlled NLI task in the zero-shot setting. We evaluate five additional models using the same prompt, next-token scoring procedure, and canonical generated set as in Section~\ref{sec:behavioral}. Each model scores the same 42,000 prompts. We focus on models of comparable scale because extending the full layer-wise causal analysis to substantially larger models would be computationally prohibitive.

Pol.\ Acc.\ is three-label accuracy on matching polarity-composition examples. The $m_i=0$ column reports $\mathrm{U}$ accuracy on the 16,000 mismatch examples. Selection Acc.\ is three-label accuracy in the multi-event experiment. Target Dir.\ measures whether changing the target polarity moves $R$ in the expected direction, Stable measures prediction invariance under distractor changes, and Ratio is the target-to-distractor mean absolute effect ratio.

\begin{table*}[t]
\centering
\caption{\textbf{Zero-shot behavioral results across models.}}
\label{tab:additional-model-screening}
\small
\setlength{\tabcolsep}{5pt}
\begin{tabular}{@{}lrrrrrr@{}}
\toprule
Model
& Pol.\ Acc.
& $m_i{=}0$ U Acc.
& Selection Acc.
& Target Dir.
& Stable
& Ratio ($\times$) \\
\midrule
Qwen3-8B
& 100.0 & 98.9 & 98.9 & 100.0 & 98.9 & 37.0 \\
Phi-4 Mini
& 93.3 & 98.9 & 89.4 & 100.0 & 92.1 & 15.3 \\
\midrule
Gemma 3 12B IT
& 99.7 & 94.8 & 98.1 & 100.0 & 98.2 & 20.1 \\
OLMo 3 7B Instruct
& 90.1 & 57.4 & 91.6 & 98.2 & 91.5 & 5.0 \\
Granite 4.1 8B
& 95.9 & 51.7 & 71.6 & 99.4 & 91.3 & 14.9 \\
Gemma 2 9B
& 50.1 & 0.1 & 50.0 & 40.6 & 100.0 & 1.1 \\
Llama 3.1 8B Instruct
& 51.9 & 0.1 & 50.0 & 75.5 & 100.0 & 2.2 \\
\bottomrule
\end{tabular}
\end{table*}

Performance varies substantially across similarly sized models. Qwen3-8B, Phi-4 Mini, and Gemma 3 reliably follow the zero-shot task, whereas OLMo 3 and Granite solve only parts of it. Gemma 2 and Llama 3.1 exhibit degenerate label behavior, predicting $\mathrm{T}$ for nearly all inputs and rarely using $\mathrm{U}$. Thus, success on the controlled task is non-trivial at this model scale. Larger models may behave differently, but evaluating their causal representations is outside our computational scope.
\section{DAS Intervention Data and Identification Controls}
\label{apx:das-controls}
\begin{table*}[t]
\centering
\caption{\textbf{Atomic DAS datasets.}
Counts are nominal generated pairs before removing relaxed
cross-split mismatch rows. Validation and test are separate splits; the mixture column gives optimization-time sampling weights.}
\label{tab:das-data}
\small
\setlength{\tabcolsep}{5pt}
\begin{tabular}{@{}lrrrrrl@{}}
\toprule
Target & Controls & Total & Train & Val. & Test &
Training mixture \\
\midrule
$p_c$
& 5 & 60,000 & 42,000 & 9,000 & 9,000
& active 80\%, inactive 20\% \\
$p_i$
& 7 & 84,000 & 58,800 & 12,600 & 12,600
& active 80\%, inactive 10\%, locality 10\% \\
$\rho_i$
& 6 & 72,000 & 50,400 & 10,800 & 10,800
& active 80\%, inactive 20\% \\
$m_i$
& 4 & 123,600 & 109,200 & 7,200 & 7,200
& transfer 80\%, anti-copy 20\% \\
\bottomrule
\end{tabular}
\end{table*}
\subsection{Dataset Sizes and Sampling}

Each DAS instance contains a base prompt, a source prompt, and a counterfactual label. The base supplies the surrounding computation, while the source supplies the target-variable value. For each target, we sample 1,000 SVO seed events from the inventory described in Appendix~\ref{apx:data_construct}.

Complete SVO tuples are split into 700 training, 150 validation, and 150 test events. Every intervention pair derived from one event remains in the same split, and training prompts cannot use held-out SVOs as distractors or mismatch events. Each event is expanded over three target-assumption positions and four assumption--claim polarity combinations.

Table~\ref{tab:das-data} reports nominal generation counts. Training emphasizes active interventions while retaining inactive, locality, and anti-copy controls. Validation and test controls are balanced independently of the optimization mixture.

For $m_i$, rows requiring relaxed held-out-event exclusion are removed after generation. The retained dataset contains 108,732 training, 7,191 validation, and 7,200 test pairs (123,123 in total). Evaluation additionally balances mismatch realizations within each polarity--control condition.
\subsection{Claim-Polarity Controls}

The $p_c$ optimization mixture is 60\% \texttt{main},
10\% \texttt{probe\_flip\_both}, 10\% \texttt{probe\_flip\_pi}, 10\% \texttt{gate\_m0}, and 10\% \texttt{label\_copy\_trap}. In \texttt{main}, the source changes $p_c$ while the open-gate base retains $p_i^{(b)}$, so the counterfactual reverses the base $\mathrm{T}/\mathrm{F}$ label.

In \texttt{probe\_flip\_both}, both source polarities are reversed. The source relation and natural label therefore equal those of the base, but transferring only $p_c^{(s)}$ must reverse the base output. In \texttt{probe\_flip\_pi}, $p_c^{(s)}=p_c^{(b)}$ while $p_i$, $\rho_i$, and the source label change, so transferring $p_c$ must preserve the base output. Together, these controls distinguish raw $p_c$ from $p_i$, $\rho_i$, and the source label.

In \texttt{gate\_m0}, both examples have $m_i=0$, and changing
$p_c$ must leave the output at $\mathrm{U}$. In \texttt{label\_copy\_trap}, the base has $m_i=0$ while the source
has $m_i=1$ and a polarized label; the required output nevertheless remains $\mathrm{U}$.

\subsection{Assumption-Polarity Controls}

The $p_i$ mixture is 30\% \texttt{main}, 30\% \texttt{active\_source\_m0}, 10\% \texttt{probe\_flip\_both}, 10\% \texttt{probe\_flip\_pc}, 5\% \texttt{gate\_m0}, 5\% \texttt{label\_copy\_trap}, and 10\% \texttt{distractor}. In \texttt{main}, $m_i=1$ and only $p_i$ changes, so the polarized output reverses. \texttt{active\_source\_m0} uses an open-gate base but obtains $p_i$ from a source with $m_i=0$ and natural label $\mathrm{U}$. The counterfactual must still follow $g(p_i^{(s)}p_c^{(b)})$, distinguishing raw $p_i$ from a gated relation or source answer.

The two identity probes are symmetric to those for $p_c$. \texttt{probe\_flip\_both} reverses both source polarities and requires the output to change under transfer of $p_i$ even though the source relation is preserved. In \texttt{probe\_flip\_pc}, $p_i^{(s)}=p_i^{(b)}$ while $p_c$, $\rho_i$, and the source label change, so the base output must remain unchanged. The two inactive controls require $\mathrm{U}$ under a closed base gate. Finally, \texttt{distractor} changes the polarity of a non-matching assumption and requires the original prediction to be preserved.

\subsection{$\rho_i$ Controls}

The $\rho_i$ mixture assigns 20\% each to
\texttt{flip\_pi}, \texttt{flip\_pc}, \texttt{hold\_both}, and \texttt{source\_m0}, and 10\% each to \texttt{gate\_m0} and \texttt{label\_copy\_trap}. Under an open base gate, changing either $p_i$ or $p_c$ reverses $\rho_i$ and must reverse the T/F output. Changing both preserves $\rho_i$ and must preserve the output. These three within-state conditions distinguish $\rho_i$ from either raw polarity.

The \texttt{source\_m0} condition uses a closed-gate source with the opposite annotated $\rho_i$. It remains part of the training mixture, but we treat it as a cross-state diagnostic rather than an identity control. The factorial audit in Section~\ref{sec:rho-context} shows that cross-state transfer can also reverse the prediction when the annotated $\rho_i$ value is unchanged. We therefore exclude \texttt{source\_m0} from confirmatory IIA and NEx. \subsection{Matching Controls}

The $m_i$ mixture assigns 40\% to \texttt{match\_to\_nomatch}, 40\% to \texttt{nomatch\_to\_match}, and 10\% to each anti-copy control. \texttt{match\_to\_nomatch} transfers $m_i=0$ into a matched base and requires $\mathrm{U}$. Conversely, \texttt{nomatch\_to\_match} transfers $m_i=1$ into an unmatched base and requires $g(\rho_i^{(b)})$. Mismatches alter the subject, verb, object, or all three components; evaluation balances these realizations.

The first anti-copy condition, \texttt{label\_copy\_trap}, uses an unmatched base and a matched source whose assumption polarity is opposite to the base polarity. Transferring only $m_i^{(s)}=1$ must produce $g(\rho_i^{(b)})$, while the source's own label is the opposite polarized label. The second condition, \texttt{label\_copy\_trap\_same\_m1}, keeps
$m_i^{(s)}=m_i^{(b)}=1$ but again reverses the source assumption polarity. Since the transferred matching value is unchanged, the target must preserve the base label even though the source label is opposite. These controls prevent a source-answer or source-polarity feature from masquerading as $m_i$.

The matching counterfactual assumes that at most one assumption event matches the claim. Under this restriction, opening or closing the designated gate uniquely determines whether its polarity relation contributes to the output.
\FloatBarrier
\section{Additional Causal Analyses}
\label{apx:causal}

\subsection{Assumption \texorpdfstring{$p_i$}{pi} subspace}
\label{apx:pi_subspace}
\begin{figure}
    \centering
    \includegraphics[width=1\linewidth]{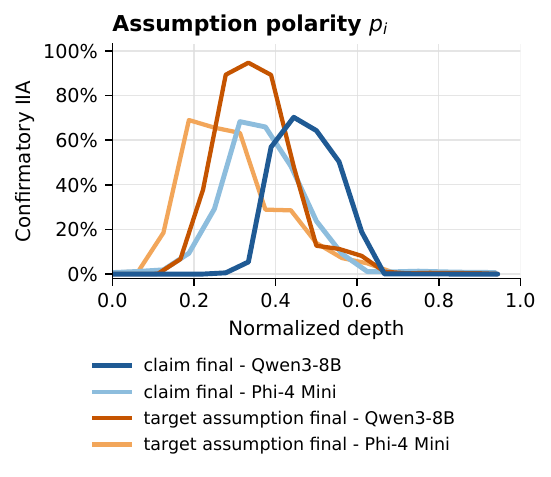}
    \caption{\textbf{Layer-wise causal identification of $p_i$} at target assumption final}
    \label{fig:pi_assumption}
\end{figure}

We report the \(p_i\) subspace identified at the target-assumption-final token in Figure~\ref{fig:pi_assumption}. In both models, assumption polarity emerges four layers earlier at this position than at the claim-final token, consistent with a subsequent transfer from the matched assumption to the claim.

\subsection{Value-Conditioned Polarity Resampling}
\label{apx:polarity-resampling}

The main-text NEx uses resampling without conditioning on the source value. We additionally separate sources carrying the same or opposite polarity as the base. The remaining high-level variables are held fixed, and source selection prefers a different lexical event. A same-value source should preserve the natural prediction, whereas an opposite-value source should produce the corresponding counterfactual. 

Table~\ref{tab:polarity-resampling} reports this diagnostic at the claim-final subspaces where raw polarity is identified. We exclude late answer-token subspaces whose polarity identity is not sufficiently confirmed.

\begin{table*}[t]
\centering
\caption{\textbf{Value-conditioned polarity resampling.}
Natural is unmodified accuracy; Same is accuracy after transferring the same polarity value. Opp.\ orig.\ and Opp.\ cf.\ evaluate an opposite-value intervention against the original and counterfactual labels, respectively.}
\label{tab:polarity-resampling}
\small
\setlength{\tabcolsep}{5pt}
\begin{tabular}{@{}llrrrr@{}}
\toprule
Target & Model & Natural & Same & Opp.\ orig. & Opp.\ cf. \\
\midrule
$p_i$ & Phi-4 Mini Instruct & 85.8 & 87.9 & 10.7 & 86.9 \\
$p_i$ & Qwen3-8B            & 96.2 & 94.7 &  6.7 & 76.6 \\
$p_c$ & Phi-4 Mini Instruct & 87.1 & 93.7 &  2.5 & 89.6 \\
$p_c$ & Qwen3-8B            & 96.2 & 97.5 &  1.8 & 94.9 \\
\bottomrule
\end{tabular}
\end{table*}
As shown in Table~\ref{tab:polarity-resampling}, transferring the same value across different lexical events preserves 87.9-97.5\% accuracy. In contrast, opposite-value transfer reduces
original-label accuracy to 1.8-10.7\% while achieving 76.6-94.9\% accuracy against the counterfactual label. Thus, the intervention systematically redirects the prediction according to
the transferred polarity rather than merely damaging model performance. This supports a transferable polarity representation that is not tied to a particular lexical event.

\subsection{Rank Robustness}
\label{apx:rank-robustness}

\begin{figure*}[p]
    \centering
    \includegraphics[width=1\linewidth]{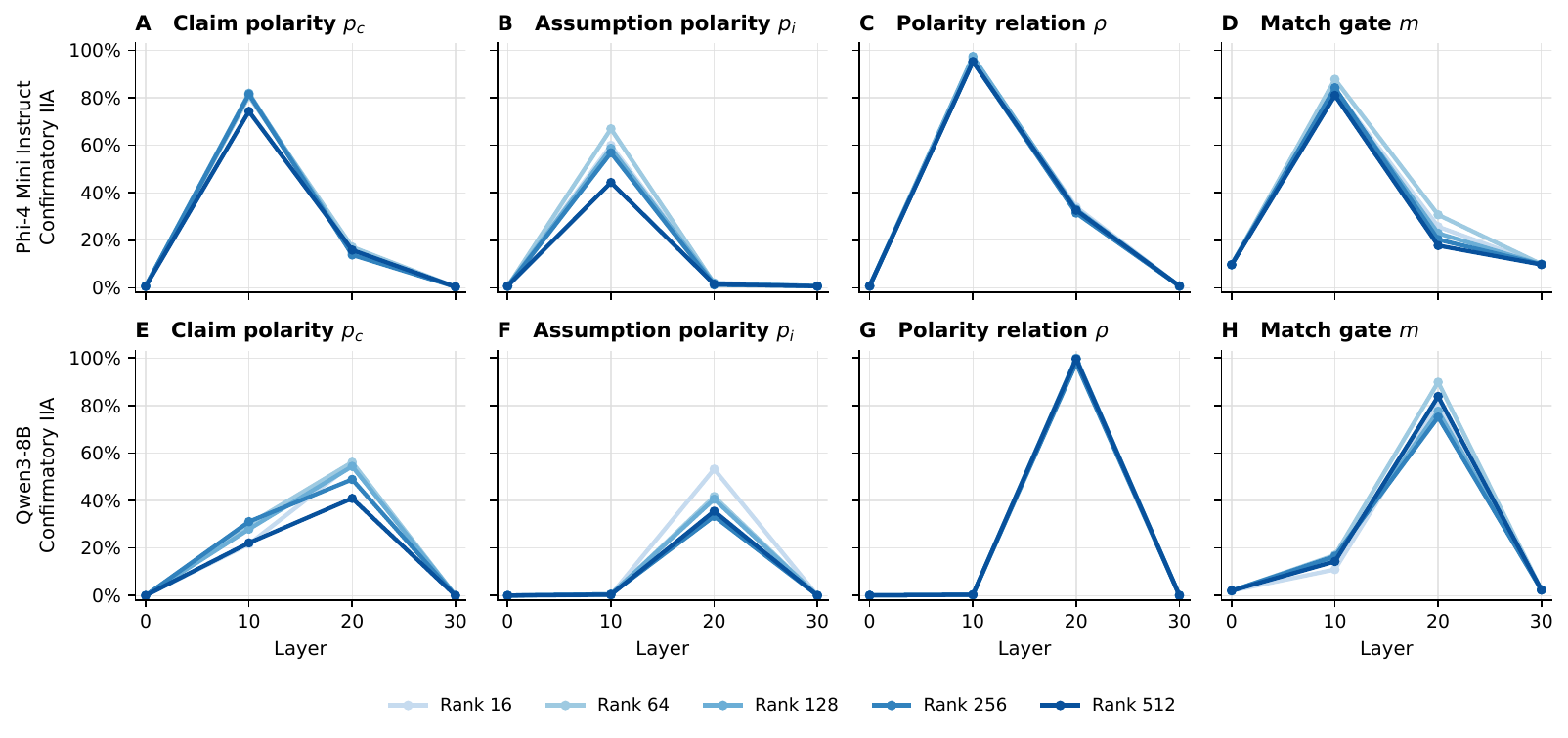}
    \caption{Rank robustness of the identified causal subspaces.}
    \label{fig:rank-robust}
\end{figure*}

Figure~\ref{fig:rank-robust} reports confirmatory IIA at the claim-final position for ranks $k \in \{16, 64, 128, 256, 512\}$ on layers 0, 10, 20, and 30. The peak layer is identical across ranks for every variable (layer 10 for Phi-4 Mini Instruct, layer 20 for Qwen3-8B). Peak IIA is nearly rank-invariant for $\rho_i$ ($\le 2.2$~pp) but varies by up to 14.7~pp for $m_i$ and 22.6~pp for $p_i$ and $p_c$, with the largest ranks generally giving the lowest peaks. $k = 64$ is within 1.1~pp of the best rank everywhere except Qwen3-8B $p_i$ (11.5~pp below $k = 16$). Peak locations remain broadly stable across ranks.

\makeatletter
\setlength{\@dblfptop}{0pt}
\setlength{\@dblfpsep}{12pt}
\setlength{\@dblfpbot}{0pt plus 1fil}
\makeatother
\section{Additional Robustness Results}
\label{apx:robustness}


\subsection{Negation-Form Results}
Each form is evaluated on 150 held-out SVO events and four polarity conditions, yielding 600 behavioral examples per form. Within-form interventions use source and base examples with the same negation form. Cross-form interventions transfer from the canonical \textit{did not} form to an alternative construction. Details in Table~\ref{tab:robust-negation-detail} and ~\ref{tab:robust-variable-detail}.
\begin{table*}[t!]
\centering
\caption{\textbf{Robustness across negation forms (\%).}
Within denotes same-form $\rho_i$ transfer; Cross transfers from a
\textit{did not} source. Claim and Answer denote the claim-final and
answer-token positions.}
\label{tab:robust-negation-detail}
\footnotesize
\setlength{\tabcolsep}{2.2pt}
\renewcommand{\arraystretch}{1.05}
\begin{tabular}{@{}lrrrrrrrrrr@{}}
\toprule
&
\multicolumn{5}{c}{Phi-4 Mini Instruct} &
\multicolumn{5}{c}{Qwen3-8B} \\
\cmidrule(lr){2-6}
\cmidrule(lr){7-11}
&
& \multicolumn{2}{c}{Within $\rho_i$}
& \multicolumn{2}{c}{Cross $\rho_i$}
&
& \multicolumn{2}{c}{Within $\rho_i$}
& \multicolumn{2}{c}{Cross $\rho_i$} \\
\cmidrule(lr){3-4}
\cmidrule(lr){5-6}
\cmidrule(lr){8-9}
\cmidrule(lr){10-11}
Negation form
& Acc. & Claim & Answer & Claim & Answer
& Acc. & Claim & Answer & Claim & Answer \\
\midrule
\textit{did not}
& 92.3 & 99.6 & 99.4 & --   & --
& 99.8 & 99.7 & 99.3 & --   & -- \\

\textit{didn't}
& 85.2 & 99.6 & 99.0 & 99.5 & 99.2
& 99.8 & 99.6 & 99.1 & 99.6 & 99.2 \\

\textit{never}
& 93.5 & 99.5 & 98.8 & 99.5 & 99.1
& 99.8 & 99.5 & 99.8 & 99.5 & 99.7 \\

\textit{did not ever}
& 98.5 & 99.5 & 99.1 & 99.6 & 99.2
& 99.8 & 99.7 & 99.7 & 99.7 & 99.6 \\

\textit{failed to}
& 86.5 & 97.4 & 96.8 & 98.4 & 98.3
& 99.8 & 98.5 & 99.5 & 99.5 & 99.5 \\

\textit{it is not the case that}
& 96.8 & 99.3 & 98.6 & 99.2 & 99.3
& 99.8 & 99.6 & 98.4 & 99.7 & 99.1 \\
\bottomrule
\end{tabular}
\end{table*}

\begin{table*}[t!]
\centering
\caption{\textbf{Variable-specific confirmatory IIA across negation
forms (\%).}
Ranges give the minimum and maximum across the six forms. The
\textit{failed to} columns report claim-final IIA.}
\label{tab:robust-variable-detail}
\small
\setlength{\tabcolsep}{6pt}
\renewcommand{\arraystretch}{1.05}
\begin{tabular}{@{}lrrrrrr@{}}
\toprule
&
\multicolumn{3}{c}{Phi-4 Mini Instruct} &
\multicolumn{3}{c}{Qwen3-8B} \\
\cmidrule(lr){2-4}
\cmidrule(lr){5-7}
Target
& Claim range & Answer range & \textit{failed to} Claim
& Claim range & Answer range & \textit{failed to} Claim \\
\midrule
$p_i$
& 50.3--68.3 & 34.5--42.0 & 50.3
& 62.1--70.2 & 7.5--15.5 & 64.1 \\

$p_c$
& 59.9--87.6 & 34.3--61.2 & 59.9
& 66.9--92.3 & 4.5--14.5 & 66.9 \\

$\rho_i$
& 97.4--99.6 & 96.8--99.4 & 97.4
& 98.5--99.7 & 98.4--99.8 & 98.5 \\

$m_i$
& 83.0--91.3 & 65.8--78.0 & 83.0
& 83.5--91.7 & 62.9--73.6 & 83.5 \\
\bottomrule
\end{tabular}
\end{table*}

\subsection{MNLI Causal Results}

Table~\ref{tab:mnli-causal-detail} expands the main-paper results with both causal positions and additional controls. Exact IIA uses the T/F/U output, whereas T/F IIA considers only the direction of the T/F preference. All causal results are averaged over three DAS seeds.

\begin{table*}[t!]
\centering
\caption{\textbf{Detailed MNLI causal results (\%).}}
\label{tab:mnli-causal-detail}
\small
\setlength{\tabcolsep}{3pt}
\renewcommand{\arraystretch}{1.03}

\begin{tabular*}{0.72\textwidth}{
    @{\extracolsep{\fill}}llrrrr@{}
}
\multicolumn{6}{c}{
    \textbf{A. Relative-polarity transfer}
} \\[3pt]
\toprule
& &
\multicolumn{2}{c}{Phi-4 Mini Instruct} &
\multicolumn{2}{c}{Qwen3-8B} \\
\cmidrule(lr){3-4}
\cmidrule(lr){5-6}
Relation & Position
& Exact IIA & T/F IIA
& Exact IIA & T/F IIA \\
\midrule
Same
& Claim final
& 61.5 & 86.6 & 81.7 & 92.8 \\
Same
& Answer token
& 60.4 & 84.2 & 81.6 & 94.6 \\
\addlinespace
Opposite
& Claim final
& 36.9 & 62.0 & 25.9 & 41.4 \\
Opposite
& Answer token
& 65.3 & 84.6 & 83.2 & 91.4 \\
\bottomrule
\end{tabular*}

\vspace{6pt}

\begin{tabular*}{0.72\textwidth}{
    @{\extracolsep{\fill}}llrrrrrr@{}
}
\multicolumn{8}{c}{
    \textbf{B. Matching transfer and controls}
} \\[3pt]
\toprule
& &
\multicolumn{3}{c}{Phi-4 Mini Instruct} &
\multicolumn{3}{c}{Qwen3-8B} \\
\cmidrule(lr){3-5}
\cmidrule(lr){6-8}
Direction & Position
& $m_i$ & $\rho_i$ & Random
& $m_i$ & $\rho_i$ & Random \\
\midrule
Polarized $\rightarrow \mathrm{U}$
& Claim final
& 55.2 & 16.5 & 4.4
& 40.7 & 8.3 & 0.9 \\
&
Answer token
& 88.9 & 5.2 & 3.9
& 77.9 & 9.9 & 1.2 \\
\addlinespace
$\mathrm{U}\rightarrow$ polarized
& Claim final
& 27.1 & 6.5 & 0.4
& 27.2 & 4.3 & 1.0 \\
&
Answer token
& 65.7 & 10.7 & 0.7
& 63.9 & 14.0 & 0.8 \\
\bottomrule
\end{tabular*}

\vspace{6pt}

\begin{tabular*}{0.72\textwidth}{
    @{\extracolsep{\fill}}lrr@{}
}
\multicolumn{3}{c}{
    \textbf{C. Claim-final polarity-contrast change after zero projection}
} \\[3pt]
\toprule
Removed subspace
& Phi-4 Mini Instruct
& Qwen3-8B \\
\midrule
Learned $\rho_i$
& $-81.0$
& $-56.1$ \\
Equal-rank random
& $-2.8$
& $-1.6$ \\
\bottomrule
\end{tabular*}

\end{table*}
At the answer token, both same- and opposite-$\rho_i$ interventions produce high T/F IIA, consistent with the transfer results reported in the main paper. Matching interventions are substantially stronger than the $\rho_i$ and random controls for movement between polarized and $\mathrm{U}$ outputs. Zeroing the learned claim-final $\rho_i$ subspace also reduces the polarity contrast much more than zeroing an equal-rank random subspace.

\end{document}